\documentclass[sigconf, screen]{acmart}
\AtBeginDocument{%
  }
\usepackage{pifont}
\usepackage{algorithm}
\usepackage{algorithmic}
\setcopyright{acmlicensed}
\copyrightyear{2018}
\acmYear{2018}
\acmDOI{XXXXXXX.XXXXXXX}
\acmConference[Conference acronym 'XX]{Make sure to enter the correct
  conference title from your rights confirmation email}{June 03--05,
  2018}{Woodstock, NY}
\acmISBN{978-1-4503-XXXX-X/2018/06}

\begin{document}

\title{ISExplore: Informative Segment Selection for Efficient Personalized 3D Talking Face Generation}

\author{Rui-Qing Sun}
\author{Ang Li}
\author{Zhijing Wu}
\authornotemark[1]
\author{Tian Lan}
\author{Qian-Yu Lu}
\author{Xing-Shan Yao}
\author{Chen Xu}
\author{Xian-Ling Mao}
\authornotemark[1]
\email{maoxl@bit.edu.cn}

% % \email{webmaster@marysville-ohio.com}
\affiliation{%
  \institution{Beijing Institute of Technology}
  \city{Beijing}
  \country{China}
}

% \affiliation{%

% \affiliation{%

% \affiliation{%

% \affiliation{%

% \affiliation{%

% \affiliation{%

% \affiliation{%

\renewcommand{\shortauthors}{Sun et al.}

\begin{abstract}

Talking Face Generation (TFG) methods based on Neural Radiance Fields (NeRF) and 3D Gaussian Splatting (3DGS) have recently achieved impressive progress in personalized talking head synthesis. However, existing methods typically require several minutes of reference video for meticulous preprocessing and fitting, resulting in hours of preparation time and limiting their practical applicability. 
In this paper, we revisit a fundamental yet underexplored question: do high-quality personalized TFG models truly require minutes-long reference videos? Our exploratory study reveals that a carefully selected reference segment of only a few seconds can often achieve performance comparable to that of using the full reference video. This finding suggests that the informativeness of reference data is more critical than its duration.
Motivated by this observation, we propose ISExplore (Informative Segment Explore), a simple yet effective segment selection strategy that automatically identifies the most informative short reference segment based on three key data quality dimensions: audio feature diversity, lip movement amplitude, and viewpoint diversity. 
Extensive experiments demonstrate that ISExplore reduces data processing and training time by over 5× for both NeRF- and 3DGS-based methods, while preserving high-fidelity generation quality. Our method provides a practical and efficient solution for personalized TFG and offers new insights into data efficiency in 3D talking face generation.
\end{abstract}



%   % sampleteaser removed for arXiv



\maketitle

\section{Introduction}
\label{sec:intro}

\begin{figure}
\centering
\includegraphics[width=0.5\textwidth]{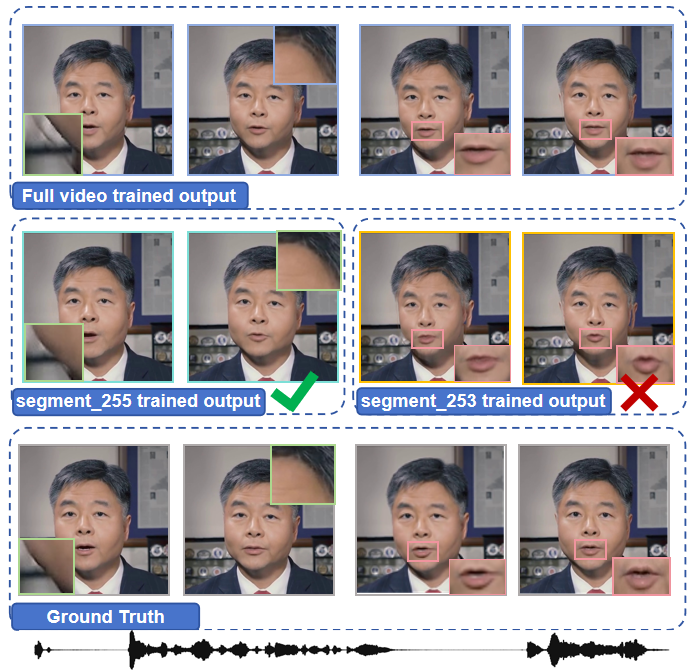}
\caption{The quality of the output videos of the models trained with different 5s segments varies greatly. 
Some segments like informative segment\_255 can achieve better results than those trained with the full video, while others have motion blur like less informative segment\_253.}
\label{fig:tfg_model}
\end{figure}

Audio-driven Talking Face Generation (TFG) has a wide range of applications in digital education, film and television production, e-commerce live streaming, and related scenarios \cite{Zhong_2023_CVPR,Jang_2024_CVPR,wang2021facevid2vid,Zhou_2020_SiggraphAsia}. Among them, \emph{personalized} TFG has attracted increasing attention due to its ability to generate high-fidelity talking videos while preserving person-specific appearance and motion characteristics \cite{guo2021adnerf,Ye_2024_NeurIPS,zhang2018LPIPS}. 

Recent advances in Neural Radiance Fields (NeRF) and 3D Gaussian Splatting (3DGS), both based on 3D representations, have significantly promoted the development of personalized TFG \cite{li2023ernerf,Tang_2025_rad-nerf,cho2024gaussiantalker,Li_2024_talkingaussian}. Unlike generalized methods \cite{ditto2025Ditto,ye2024real3dportrait}, these personalized 3D representation-based approaches offer superior performance in maintaining identity consistency, preserving facial details, and accurately modeling speaker-specific lip motions \cite{Ye_2024_NeurIPS,Peng_2024_synctalk,Li_2025_instag,Ye_2024_geneface}. In general, these methods first capture and encode the 3D information of a target individual from reference videos, and then learn a personalized mapping from audio to facial dynamics. As a result, even under the same driving audio, different individuals can still exhibit distinct personalized motion patterns. \textbf{However}, for most existing methods \cite{Peng_2024_synctalk,Bi_2024_NeRF-AD,li2023ernerf,cho2024gaussiantalker,Li_2024_talkingaussian}, a sufficiently long reference video (typically 3--5 minutes) is required to ensure adequate 3D information for stable modeling. Consequently, for each new target individual, it often takes hours to preprocess several minutes of reference video and train a new model, which substantially limits the practical applicability of personalized TFG.

This raises a fundamental question: \emph{do personalized TFG models truly require minutes-long reference videos for each target individual?} To answer this question, we conduct exploratory case studies by training radiance field-based models from scratch using short reference segments. Our study reveals two key findings. First, a carefully selected reference segment of only a few seconds can still yield high-quality output comparable to that obtained using the full reference video. Second, not all short segments are equally informative: some lead to stable and high-fidelity synthesis, while others produce jittery motion and degraded visual quality, as illustrated in Figure~\ref{fig:tfg_model} (e.g., segment 255 versus segment 253). These observations suggest that previous methods may spend substantial computation fitting long but only partially useful video content. More importantly, they indicate that, for personalized TFG, the \emph{informativeness} of the reference video matters more than its duration. Therefore, it is crucial to develop an automatic strategy for identifying informative short reference segments, enabling efficient yet high-quality personalized TFG training.

Motivated by this observation, we first conduct extensive preliminary studies to analyze the characteristics of reference segments that lead to high-quality outputs. Based on these findings, we propose a simple yet effective data selection strategy, termed \textbf{ISExplore} (\textbf{In}formative \textbf{S}egment \textbf{Explore}), for automatically identifying informative short reference segments. Specifically, our strategy considers three key data quality dimensions that are critical for informative segment discovery:
1) \textbf{audio feature diversity}, which reflects the richness of speech content and is analyzed under the known interdependence between audio and camera-view dynamics \cite{Bi_2024_NeRF-AD,Peng_2023_emotalk,Zhou_2019_DAVS};
2) \textbf{lip movement amplitude}, which characterizes the motion intensity of mouth movements in the frequency domain via Fourier analysis; and
3) \textbf{viewpoint diversity}, which measures the complexity of head and camera motion.
By jointly leveraging these cues, ISExplore filters out low-value reference content with minimal additional overhead, thereby reducing both data preparation cost and downstream training time. Unlike prior approaches that mainly focus on reducing model training cost by introducing additional priors or general information \cite{Li_2025_instag}, our method targets an earlier but equally critical stage: \emph{reference video selection before training}.

Extensive experiments on representative personalized TFG benchmarks \cite{Peng_2024_synctalk,Li_2025_instag,Ye_2024_geneface} demonstrate the effectiveness of ISExplore. Specifically, both NeRF- and 3DGS-based models trained with ISExplore using only 5 seconds of reference video achieve performance that is comparable to, or even better than, state-of-the-art (SOTA) TFG models trained on reference videos that are more than 10$\times$ longer. Even after accounting for the runtime of ISExplore itself, our method reduces the total data processing and training time by over 5$\times$, while still maintaining high-fidelity output quality. Furthermore, both theoretical analysis and empirical results show that our method is compatible with previous approaches that incorporate additional priors, making it a practical and complementary plug-in for existing personalized TFG pipelines.

Our contributions are summarized as follows:
\begin{itemize}
    \item We show that high-quality personalized TFG models can be trained from scratch using only a few seconds of \emph{informative} reference video.
    \item We propose ISExplore, a simple yet effective reference segment selection strategy that improves the data preprocessing and training efficiency of personalized TFG models.
    \item Extensive experiments demonstrate that ISExplore can be seamlessly integrated into mainstream NeRF- and 3DGS-based personalized TFG frameworks, achieving competitive generation quality while significantly reducing computational cost.
\end{itemize}

\section{Related Work}

\subsection{Personalized Talking Face Generation}

Personalized Talking Face Generation (TFG) aims to synthesize realistic talking portraits that preserve the identity and motion characteristics of a specific individual. Early methods mainly relied on 2D generative models, but often struggled to produce natural head movement and realistic facial dynamics \cite{chen2019_early_method_HCTFG, chen2018_lipGan, prajwal2020_WAV2LIP}. 

Recent progress in 3D representations, especially Neural Radiance Fields (NeRF) \cite{Athar_2022_RigNeRF, Bi_2024_NeRF-AD, guo2021adnerf} and 3D Gaussian Splatting (3DGS) \cite{cho2024gaussiantalker, Li_2024_talkingaussian, Li_2025_instag}, has substantially advanced personalized TFG. Compared with generalized methods \cite{ditto2025Ditto,ye2024real3dportrait}, these approaches build subject-specific 3D representations from reference videos and learn personalized audio-driven facial motion, leading to stronger identity consistency and more accurate lip synchronization.

Representative NeRF-based methods include RAD-NeRF \cite{Tang_2025_rad-nerf}, ER-NeRF \cite{li2023ernerf}, GeneFace \cite{Ye_2024_geneface}, and SyncTalk \cite{Peng_2024_synctalk}, while 3DGS-based methods such as TalkingGaussian \cite{Li_2024_talkingaussian}, GaussianTalker \cite{cho2024gaussiantalker}, and InstaG \cite{Li_2025_instag} improve rendering sharpness and visual quality.

Despite these advances, \emph{efficiency} remains a major bottleneck. Most personalized TFG methods still require several minutes of reference video for preprocessing, fitting, and subject-specific training. Existing acceleration efforts mainly focus on the \emph{model side}, such as LoRA-based adaptation \cite{ye2024mimictalk,hu2022lora} or pre-trained motion priors \cite{Li_2025_instag}. In contrast, our work is \emph{data-centric}: instead of modifying the model or introducing additional priors, we study how to automatically identify the most informative short reference segment for efficient personalized TFG training.

\subsection{Data Selection for Efficient Training}

Data selection has recently emerged as an effective paradigm for improving training efficiency in machine learning. In particular, large language models (LLMs) are often pre-trained on massive corpora and then adapted to downstream tasks using carefully selected data. Prior studies have shown that high-quality and informative training subsets can substantially reduce training cost while maintaining strong performance \cite{chung2024_scaling_IFLM,wang2023selfinstruct,beeching2023stackllama,koepf2023_openassistant_conversations,zhou2023_LIMA,liu2024_DEITA,li2025_LIMR,ye2025_LIMO}. For instance, LIMA \cite{zhou2023_LIMA} demonstrated that fine-tuning with only 1,000 carefully curated instruction-response pairs can achieve competitive alignment performance. DEITA \cite{liu2024_DEITA} further analyzed the characteristics of ``good data,'' while subsequent works such as LIMR \cite{li2025_LIMR} and LIMO \cite{ye2025_LIMO} reinforced the effectiveness of data-efficient learning.

However, unlike text data, personalized TFG relies on \emph{multi-modal temporal video data}, where the quality of a reference segment depends not only on semantic content, but also on audio diversity, facial motion patterns, and viewpoint changes. As a result, the notion of ``good data'' in personalized TFG is substantially more complex than in text-based learning. To the best of our knowledge, existing personalized TFG methods have not systematically studied \emph{which parts of a reference video are actually most informative for training}. 

Our work fills this gap by introducing a simple yet effective reference segment selection strategy tailored for personalized TFG. Rather than treating all reference frames as equally useful, we explicitly model the informativeness of short video segments and show that selecting a few high-value seconds can significantly improve training efficiency without sacrificing generation quality.
% In this paper, we explore the selection strategy of informative segments from several minutes of reference videos for effective personalized TFG models training.

\section{Preliminary Study: What Makes a Reference Segment Informative?\label{sec:Preliminary_Study}}

In this section, we investigate a central question of this work: \emph{what makes a short reference segment informative for personalized TFG training?} To answer this question, we first formulate the problem from a data selection perspective, and then conduct a series of preliminary studies to analyze which properties of reference segments are most related to final synthesis quality. In particular, we focus on three factors: \textbf{audio feature diversity}, \textbf{lip movement amplitude}, and \textbf{camera-view complexity}. These analyses provide the empirical basis for the segment selection strategy proposed in the next section.

\subsection{Task Formulation: Data Selection\label{sec:pre_study_task_formulation}}

To investigate the relationship between reference video characteristics and final output quality, we formulate the problem within a data selection framework. Specifically, given a multi-minute reference video, we divide it into equal-length segments of duration $m$ seconds, yielding a candidate set
\[
X = \{x_1, x_2, \dots, x_n\},
\]
where each $x_i$ denotes an individual $m$-second reference segment. Our goal is to identify a segment $x_s \in X$ such that a personalized TFG model trained \emph{from scratch} on $x_s$ can produce high-fidelity output.

Let $\pi$ denote a segment selection strategy, and let $Q(x_s)$ denote the quality of the output video generated by the model trained on segment $x_s$. Then, the optimal strategy $\pi^*$ can be formulated as selecting the segment that maximizes the final output quality:
\begin{equation}
\pi^* = \arg \max Q(x_s).
\end{equation}

This formulation highlights the core challenge of our work: rather than treating all parts of a reference video as equally useful, we aim to identify the \emph{most informative} segment for efficient and high-quality personalized TFG training.

\subsection{Experimental Setup
\label{sec:pre_study_exp_setup}}

We conduct experiments on a single RTX A6000 GPU to analyze how different properties of short reference videos affect the final output quality of personalized TFG models. Based on the design of mainstream NeRF- and 3DGS-based TFG methods, we focus on three reference-segment factors: 1) \textbf{audio feature diversity}, 2) \textbf{lip movement amplitude}, and 3) \textbf{camera-view complexity}.

Both NeRF- and 3DGS-based personalized TFG models take embedded audio features and camera poses as inputs, and their effects on the generated frames are inherently coupled \cite{Bi_2024_NeRF-AD,Li_2024_talkingaussian,Zhou_2019_DAVS}. In practice, when models are trained with several minutes of reference video, this redundancy often helps decouple these factors and leads to accurate lip-shape modeling. However, when the reference video is reduced to only a few seconds, the influence of each factor becomes much more critical. Therefore, to better understand their individual effects, we construct controlled simulation settings and analyze them separately. More details are in the Appendix.

\begin{figure}[h]
\centering
\includegraphics[width=0.5\textwidth]{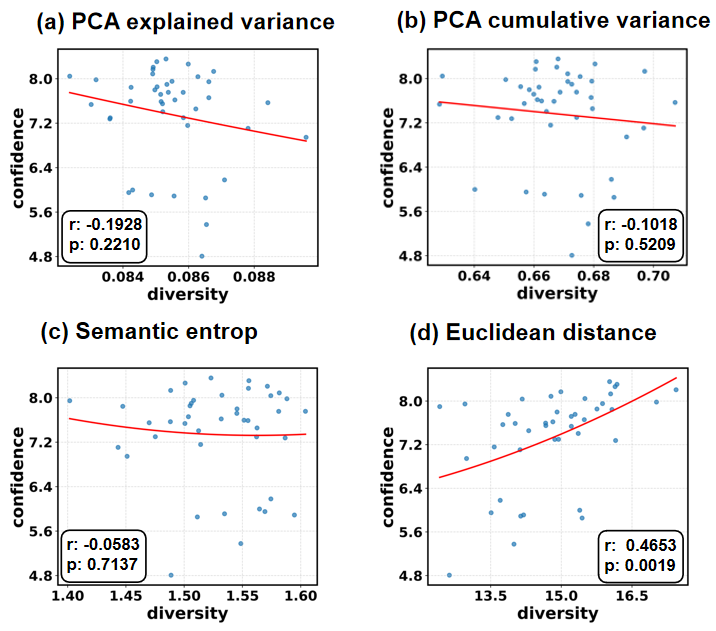}
\caption{Relationship between different audio feature diversity measures and lip synchronization confidence (Sync-C).}
\label{fig:diversity_analysis}
\end{figure}

\subsection{Audio Feature Diversity \label{sec:Audio_Feature_Diversity}}

\textbf{Motivation.}
Previous studies have shown that high-quality 3D reconstruction of static objects can often be achieved with only a limited number of images \cite{Zhang_2024_CoR-GS,Shi_2024_ZeroRF}. This naturally raises the question: \emph{can a personalized TFG model also learn reliable lip-audio correspondence from only a few seconds of reference audio?} If so, what type of short audio segment is most beneficial?

\textbf{Metric design.}
To characterize the diversity of audio information in a reference segment, we evaluate four candidate measures:
1) the average explained variance of the largest principal component,
2) the average cumulative variance of the largest principal component,
3) semantic entropy \cite{Han_2022_SE——indicator}, and
4) the average Euclidean distance between audio feature vectors.
Specifically, we first extract frame-level audio embeddings using the AVE encoder \cite{Peng_2024_synctalk}, and then compute the corresponding diversity statistics using the four measures above.

\textbf{Analysis.}
As shown in Figure~\ref{fig:diversity_analysis}, we plot the relationship between each diversity measure and the final lip synchronization quality. After removing outliers, we fit the relationship using quadratic regression. Among all candidates, the average Euclidean distance between feature vectors exhibits the clearest positive correlation with output quality, with a $p$-value of 0.0019 and an $R^2$ score of 0.4653. This result suggests that reference segments with richer and more diverse audio content tend to provide more useful supervisory signals for learning audio-to-lip mappings. Therefore, in the subsequent sections, we use the average Euclidean distance between AVE feature vectors as our measure of \emph{audio feature diversity}.

\begin{figure}[htbp]
\centering
\includegraphics[width=0.5\textwidth]{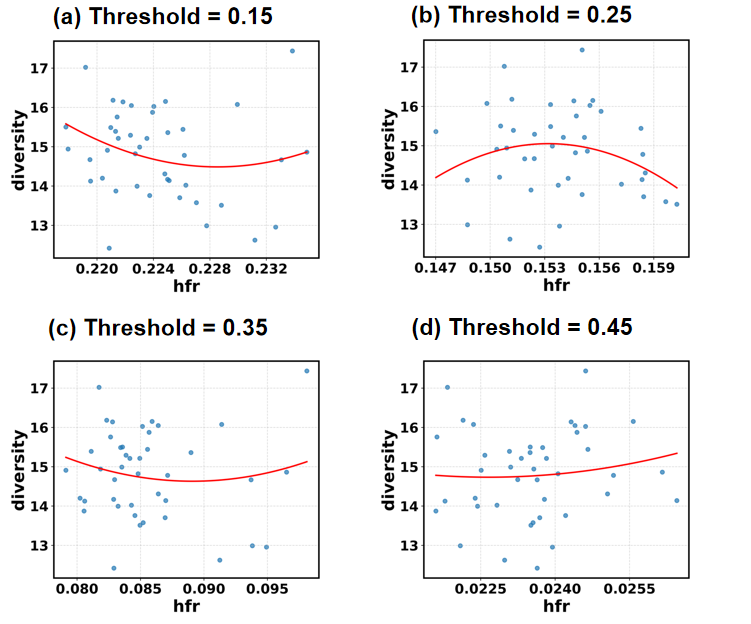}
\caption{Relationship between audio feature diversity and the \textbf{H}igh-\textbf{F}requency information \textbf{R}atio (\textbf{HFR}) of lip motion under different thresholds.}
\label{fig:frq_analysis}
\end{figure}

\subsection{Lip Movement Amplitude \label{sec:Lip_movement_amplitude}}

\textbf{Motivation.}
Previous studies have shown that neural networks tend to fit low-frequency information more easily during training \cite{Fridovich-Keil_2022_special_bias,Xu_2020_F-principle,Rahaman_2019_FQbias}. This suggests that, for reference segments of the same duration, lip motion with lower temporal complexity may be easier for personalized TFG models to fit and generalize from. Therefore, we analyze the temporal frequency characteristics of lip motion to quantify its movement amplitude and complexity.

\textbf{Metric design.}
We model lip motion as the temporal evolution of 3DMM landmark coordinates \cite{Paysan_2009_3DMM}. Specifically, we treat the continuous motion trajectory of each lip landmark as a discrete temporal signal and apply the discrete Fourier transform (DFT) to obtain its spectral distribution. Let
\[
\hat{X} = \mathcal{F}(x_{48:68}),
\]
where $\hat{X}$ denotes the frequency-domain representation after the DFT, $x_{48:68}$ represents the lip landmark coordinates (points 48--68), and $\mathcal{F}$ is the Fourier transform operator. We then define the \textbf{High-Frequency Ratio (HFR)} as:
\begin{equation}
\text{HFR} = \frac{\sum_{i \in \text{high}} |\hat{X}_i|}{\sum_{i=1}^{N} |\hat{X}_i|}
\label{HFR}
\end{equation}
where $\hat{X}_i$ is the $i$-th frequency component, and the numerator sums the components whose frequencies exceed a threshold $T$.

\textbf{Analysis.}
To examine whether lip movement complexity is simply correlated with audio diversity, we plot the relationship between HFR and audio feature diversity under thresholds ranging from 0.15 to 0.45, as shown in Figure~\ref{fig:frq_analysis}. The results show no clear linear relationship, indicating that lip movement amplitude and audio diversity capture different aspects of segment informativeness and should be treated as complementary factors. We further observe that when the threshold is set to 0.25, the samples are more evenly distributed, leading to a more stable distinction between segments. Therefore, we empirically set $T=0.25$ in the subsequent experiments.

It is worth noting that, even under simulation settings, it is difficult to construct reference videos with identical audio and pose conditions but systematically varied lip-motion frequency characteristics. Therefore, in later ablation studies, we directly validate the contribution of lip movement amplitude on real data, as shown in Table~\ref{ab_exp}.

\begin{table}[htbp]
\centering
\begin{tabular}{c|c|c}
\toprule
Camera Poses & Sync-C & Sync-D \\
\midrule
SyncTalk (5 poses)  & 8.55 & 6.61 \\
SyncTalk (10 poses) & 8.40 & 6.68 \\
SyncTalk (20 poses) & 8.24 & 6.82 \\
SyncTalk (50 poses) & 8.19 & 6.75 \\
\bottomrule
\end{tabular}
\caption{Impact of different numbers of camera views on output video quality.}
\label{pose_effect}
\end{table}

\subsection{The Impact of Camera Views \label{sec:The_impact_of_camera_views}}

\textbf{Motivation.}
Previous personalized TFG methods generally assume that facial motion (especially lip motion) should be primarily driven by audio, while global head motion should be independently controlled by camera pose \cite{Bi_2024_NeRF-AD,guo2021adnerf,Li_2024_talkingaussian}. However, when a short reference segment contains highly dynamic or diverse camera views, these two factors may become strongly entangled, making it more difficult for the model to disentangle audio-driven lip motion from pose-driven appearance changes.

\textbf{Metric design.}
To capture this effect, we use the \textbf{number of camera views} within a reference segment as a simple proxy for camera-view complexity. Intuitively, a larger number of camera viewpoints implies more head-pose variation and stronger coupling between pose and speech motion.

\textbf{Analysis.}
As shown in Table~\ref{pose_effect}, the output quality gradually decreases as the number of camera views increases. This suggests that, when the available reference video is short, simpler and more focused head motion helps the model learn more stable and accurate personalized dynamics. This observation is further validated by the ablation experiments on real data reported later in Table~\ref{ab_exp}.

\begin{figure*}[htbp]
\centering
\includegraphics[width=\textwidth]{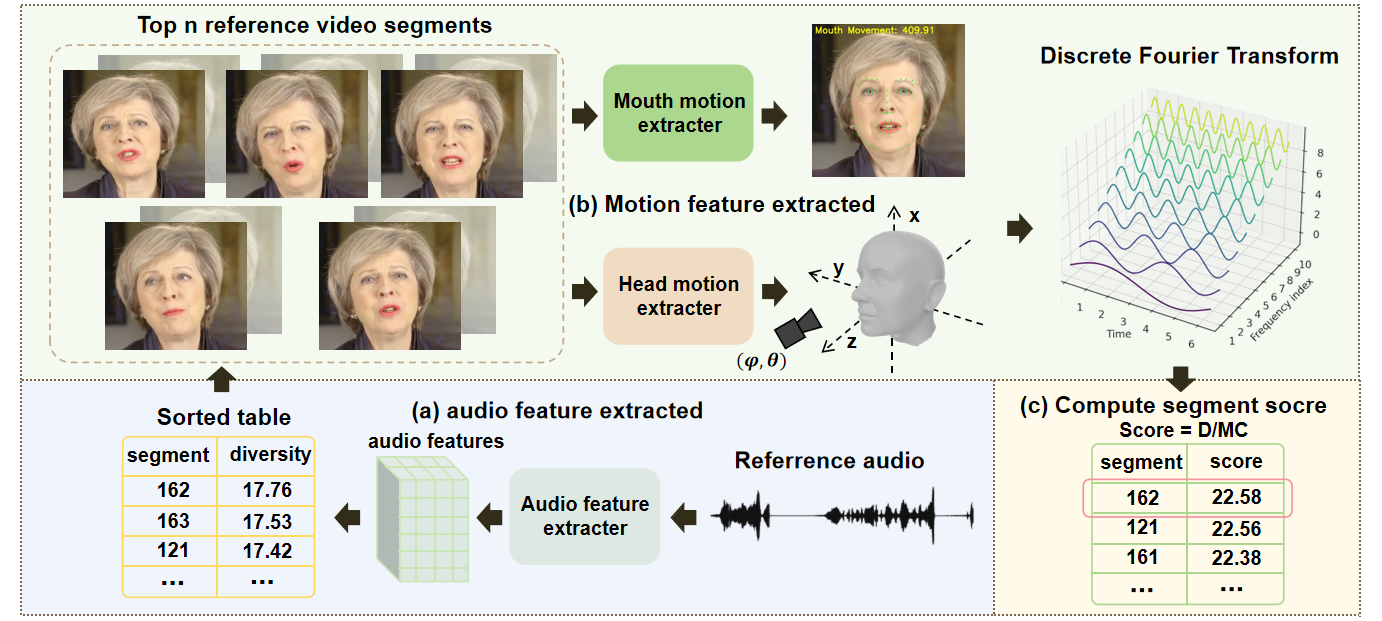}
\caption{Overview of the ISExplore pipeline. (a) Candidate reference segments are first ranked by audio diversity. (b) For the top-ranked candidates, motion complexity is estimated via landmark-based decomposition and frequency analysis. (c) Final informativeness scores are computed to select the most informative segment.}
\label{fig:isx_stratefy}
\end{figure*}

\section{ISExplore: Informative Segment Explore}

Based on the preliminary findings in Section~\ref{sec:Preliminary_Study}, we propose \textbf{ISExplore}, a simple yet effective reference segment selection strategy for efficient personalized TFG training. The key idea is to identify short reference segments that are both \emph{informative} and \emph{easy to fit}. Specifically, an ideal segment should contain diverse audio content while avoiding overly complex motion patterns.

As illustrated in Figure~\ref{fig:isx_stratefy}, ISExplore follows a three-stage pipeline, which is fully consistent with Algorithm~\ref{fis:algorithm}:  
a) rank candidate segments by audio diversity,  
b) estimate motion complexity for the top-$m$ candidates, and  
c) compute the final informativeness score to select the best segment. 
The informativeness score $I$ is defined as:
\begin{equation}
    I = \frac{D}{MC + 1 \times 10^{-8}},
\label{I_score}
\end{equation}
where $D$ denotes audio diversity and $MC$ denotes motion complexity. A larger $I$ indicates a better trade-off between information richness and fitting difficulty. The following are detailed explanations.

\textbf{Step 1: Audio Diversity Ranking}. 
Corresponding to Figure~\ref{fig:isx_stratefy}(a), we first construct a candidate set $X$ by extracting fixed-length segments from the reference video using a sliding window with a stride of 1 second (Algorithm~\ref{fis:algorithm}, line 1):
\[
X = \{x_1, x_2, \dots, x_N\}.
\]

We then compute the audio diversity score $D$ for each segment using the metric introduced in Section~\ref{sec:Audio_Feature_Diversity}, and sort all candidates in descending order (Algorithm~\ref{fis:algorithm}, lines 2--3), obtaining:
\[
X^* = \{x_1^*, x_2^*, \dots, x_N^*\}.
\]

To improve efficiency, we retain only the top-$m$ candidates with the highest audio diversity for further evaluation.

\textbf{Step 2: Motion Complexity Estimation}.
Corresponding to Figure~\ref{fig:isx_stratefy}(b), we estimate motion complexity from landmark trajectories for each candidate in $X_{(m)}^*$ (Algorithm~\ref{fis:algorithm}, lines 5--7).

Specifically, we extract 3DMM lip landmarks and jointly model two types of motion:
\begin{itemize}
    \item \textbf{Local lip deformation}, corresponding to the \emph{mouth motion extractor}, which captures articulation such as mouth opening and closing;
    \item \textbf{Global landmark displacement}, corresponding to the \emph{head motion extractor}, which reflects head motion and camera-view variation.
\end{itemize}

The resulting landmark trajectories are then transformed into the frequency domain via DFT, as shown in Figure~\ref{fig:isx_stratefy}(b), and the high-frequency ratio in Eq.~(\ref{HFR}) is used as the motion complexity score $MC$.

\textbf{Step 3: Informativeness-Based Selection}.
Corresponding to Figure~\ref{fig:isx_stratefy}(c), we compute the informativeness score $I$ for each candidate (Algorithm~\ref{fis:algorithm}, lines 8--9), and select the segment with the highest score as the final informative reference segment (lines 10--12):
\[
x_{\pi_{\text{ISExplore}}} = \arg\max_{x_i \in X_{(m)}^*} I(x_i).
\]

In all subsequent experiments, we use $m=5$ or $m=10$ as representative settings to demonstrate the effectiveness of ISExplore.

\begin{algorithm}[h]
\caption{ISExplore: Informative Reference Segment Selection}
\label{fis:algorithm}
\textbf{Input}: Complete reference video, segment length $n$, number of retained candidates $m$ \\
\textbf{Output}: Selected informative segment $x_{\pi_{\text{ISExplore}}}$
\begin{algorithmic}[1]
\STATE Construct candidate set $X$ by extracting $n$-second segments from the reference video with a stride of 1 second
\STATE Rank all segments in $X$ according to audio diversity $D$, and obtain sorted set $X^*$
\STATE Retain the top-$m$ segments to form $X_{(m)}^*$
\STATE Initialize an empty score set $I^*$
\FOR{each segment $x_i^* \in X_{(m)}^*$}
    \STATE Extract 3DMM lip landmarks $LM(x_i^*)$
    \STATE Transform landmark trajectories into the frequency domain
    \STATE Estimate motion complexity $MC(x_i^*)$ using Eq.~(\ref{HFR})
    \STATE Compute informativeness score $I(x_i^*) = D(x_i^*) / (MC(x_i^*) + 1 \times 10^{-8})$
    \STATE Add $I(x_i^*)$ to $I^*$
\ENDFOR
\STATE Sort $I^*$ and select the segment with the highest score
\STATE \textbf{return} $x_{\pi_{\text{ISExplore}}}$
\end{algorithmic}
\end{algorithm}

\section{Experiments}

\subsection{Experimental Setup}

\textbf{Dataset and Implementation Details}.
In our main experiments, to ensure a fair evaluation under practical computational constraints, we collect and curate the evaluation videos used in previous personalized TFG studies \cite{Ye_2024_geneface,Peng_2024_synctalk,li2023ernerf,Tang_2025_rad-nerf}, forming a non-overlapping subset of the HDTF \cite{zhang2021HDTF} dataset following the standard experimental protocol described in \cite{Li_2025_instag}. The selected videos cover diverse identities and speaking styles, with no identity overlap. The average frame rate is approximately 25 fps, and each frame contains a centrally aligned portrait. What's more, to further validate the effectiveness of our method, we additionally select 40 random subjects from the HDTF dataset. For each subject, the reference video is truncated to its first 3 minutes to construct a controlled yet realistic evaluation setting.

For audio feature extraction, we adopt the AVE module \cite{Peng_2024_synctalk} within the ISExplore strategy, as it has been widely used in previous personalized TFG methods and has demonstrated strong performance \cite{Li_2025_instag,Peng_2024_synctalk,Ye_2024_geneface}. More implementation details are provided in the Appendix.

\textbf{Baselines}.
To compare with state-of-the-art (SOTA) methods, we select representative personalized TFG models from the two dominant 3D representation paradigms: NeRF-based methods, including ER-NeRF \cite{li2023ernerf}, RAD-NeRF \cite{Tang_2025_rad-nerf}, and SyncTalk \cite{Peng_2024_synctalk}; and 3DGS-based methods, including GaussianTalker \cite{cho2024gaussiantalker}, TalkingGaussian \cite{Li_2024_talkingaussian}, and InstaG \cite{Li_2025_instag}.
Following prior work \cite{Peng_2024_synctalk,Tang_2025_rad-nerf}, we use the final 1/11 portion of each video as the test set.
To further provide a broader comparison, we additionally include recent one-shot talking portrait generation methods based on a single reference image, including Ditto \cite{ditto2025Ditto} and Real3D-Portrait \cite{ye2024real3dportrait}.

\textbf{Metrics}.
We evaluate rendering quality using PSNR and SSIM \cite{Wang_2004_ssim}. Following prior work \cite{Li_2025_instag,Peng_2024_synctalk,ye2024mimictalk}, we additionally report NIQE \cite{mittal2013NIQE}, BRISQUE \cite{mittal2012BRISQUE}, LPIPS \cite{zhang2018LPIPS}, and FID \cite{heusel2017FID} for perceptual quality assessment.
For motion quality, we use SyncNet confidence (Sync-C) and SyncNet distance (Sync-D) \cite{Prajwal_2020_sync_c_d} to evaluate lip synchronization accuracy. In addition, to highlight the efficiency advantage of our method, we report both \textbf{data processing time} and \textbf{training time} for adapting each model to a new target subject.

\subsection{Main Results}

\begin{table*}[t]
\centering
\begin{tabular}{c|cccccc}
\toprule
{\textbf{Methods}} & \multicolumn{2}{c}{\textbf{Rendering Quality}} & \multicolumn{2}{c}
{\textbf{Motion Quality}} & \multicolumn{2}{c}{\textbf{Efficiency (Time Cost)}} \\
& PSNR $\uparrow$ & SSIM  $\uparrow$ & Sync-C $\uparrow$ & Sync-D $\downarrow$ & Processing $\downarrow$ & Training $\downarrow$ \\ 
\midrule
ER-NeRF \cite{li2023ernerf} & 25.17 & 0.88 & 5.41 & 9.06 & 1h 20min & 2h 32min \\ 
RAD-NeRF \cite{Tang_2025_rad-nerf} & 28.82 &0.91 & 5.07 & 8.73 & 1h 32min & 3h 48min \\  
GaussianTalker \cite{cho2024gaussiantalker}  & 29.82 & 0.92 & 5.40 & 9.28 & 57min & 3h 7min\\ 
TalkingGaussian \cite{Li_2024_talkingaussian}  & 29.59 & 0.91 & 5.37 & 9.04 & 1h 20min & 48min \\ 
Ditto \cite{ditto2025Ditto}  & 29.79 & 0.92 & 5.46 &  9.35 & - & -\\ 
Real3dportrait \cite{ye2024real3dportrait}  & 25.70 & 0.85 & 6.12 & 8.38 & - & - \\ 
\midrule
Instag \cite{Li_2025_instag}  & 28.17 & 0.90 & 5.44 & 9.00 & 11min & 12min \\ 
Instag (with ours 5s)  & 29.16 & 0.91 & 6.00 & 8.46 & 12min & 12min \\ 
\midrule
Synctalk \cite{Peng_2024_synctalk}   & 30.82 & 0.94 & 7.33 & 7.36 & 1h 10min & 2h \\
Synctalk (with ours 5s) & 30.51 & 0.94 & 6.00 & 8.97 & 9min (\textbf{$\times$ 7.77}) & 20min (\textbf{$\times$ 6}) \\
Synctalk (with ours 10s) & 30.22 & 0.94 & 6.30 & 8.14 & 12.5min (\textbf{$\times$ 5.6})& 20min (\textbf{$\times$ 6}) \\ 
Synctalk (with ours 10s*) & 30.14 & 0.94 & 6.64 & 8.03 & 12.5min (\textbf{$\times$ 5.6})& 36min (\textbf{$\times$ 3.3}) \\ 
\bottomrule
\end{tabular}
\caption{Quantitative results. Our method significantly reduces the data processing and training time required for training personalized TFG models from scratch, while remaining compatible with methods that introduce additional prior knowledge. One-shot methods require no extra data processing or training time during inference.}
\label{tab:method_comparsion}
\end{table*}
\textbf{Quantitative Results}.
Table~\ref{tab:method_comparsion} reports the quantitative comparison with SOTA methods.

For 3DGS-based models, replacing InstaG's original random segment selection with ISExplore consistently improves both rendering and lip-sync quality, indicating that the selected segments are more informative for training.

For NeRF-based models, although SyncTalk trained on full-length videos achieves the best overall quality, our selected short segments still yield competitive rendering and synchronization performance while reducing the total computational cost to less than one-sixth of the original setting.

We further observe that 10-second segments consistently outperform 5-second ones, suggesting that while short clips are already sufficient for effective personalized training, segment informativeness remains crucial to final performance.

\textbf{Additional Perceptual Results}.
\begin{table*}[h]
\centering
\begin{tabular}{c|cccccc}
\toprule
{\textbf{Methods}}
& NIQE$\downarrow$ 
& BRISQUE $\downarrow$ 
& LPIPS $\downarrow$ & FID $\downarrow$ & Processing $\downarrow$& Training $\downarrow$\\ 
\midrule
ER-NeRF \cite{li2023ernerf} & 24.053 & 47.818 & 0.063 & 15.510 & 1h20min& 2h32min\\ 
RAD-NeRF \cite{Tang_2025_rad-nerf} & 24.068 &50.369 & 0.058 & 9.872 & 1h32min& 3h48min\\  
GaussianTalker \cite{cho2024gaussiantalker}  & 24.351 & 48.460 & 0.043 & 8.422 & 57min& 3h7min\\ 
TalkingGaussian \cite{Li_2024_talkingaussian}  & 25.128 & 47.385 & 0.040 & 8.051 & 1h20min& 48min\\
Ditto \cite{ditto2025Ditto}& \textbf{23.407}& 48.047& \textbf{0.029}& 17.457& -&-\\
Real3dportrait \cite{ye2024real3dportrait}&25.135 & 51.755& 0.113&23.124 & - & -\\ 
\midrule
Instag (Random)  & 24.748 & 47.601 & 0.048 & 9.466 & 11min& 12min\\ 
Instag (with ours 5s)  & 24.615 & 47.149 & 0.042 & 9.742 & 12min& 12min\\ 
\midrule
Synctalk \cite{Peng_2024_synctalk}   & 23.921 & 41.709 & \underline{0.030} & \underline{6.335} & 1h10min& 2h\\
Synctalk (with ours 5s) & 23.842  & 41.947 & 0.031 & 8.105 & 9min (\textbf{$\times$ 7.77}) & 20min (\textbf{$\times$ 6})\\
Synctalk (with ours 10s) & 23.645 & \textbf{41.145} & \underline{0.030} & 6.492 & 12.5min (\textbf{$\times$ 5.6})& 20min (\textbf{$\times$ 5.6})\\ 
Synctalk (with ours 10s*) & \underline{23.640}& \underline{41.661} & \underline{0.030} & \textbf{5.744} & 12.5min (\textbf{$\times$ 5.6})& 36min (\textbf{$\times$ 3.3})\\ 
\bottomrule
\end{tabular}
\caption{Additional quantitative results from perceptual quality perspectives. The best result is highlighted in bold, and the second-best result is underlined.}
\label{tab:method_comparsion_others}
\end{table*}
Table~\ref{tab:method_comparsion_others} further reports NIQE, BRISQUE, LPIPS, and FID. Notably, SyncTalk trained on 10-second segments selected by ISExplore achieves perceptual quality comparable to, or in some cases even better than, full-length training. In particular, the 10s* setting achieves the best FID while requiring substantially lower computational cost. These results further support our central claim: the informativeness of the reference segment is more critical than its duration.

\begin{figure*}[h]
\centering
\includegraphics[width=\textwidth]{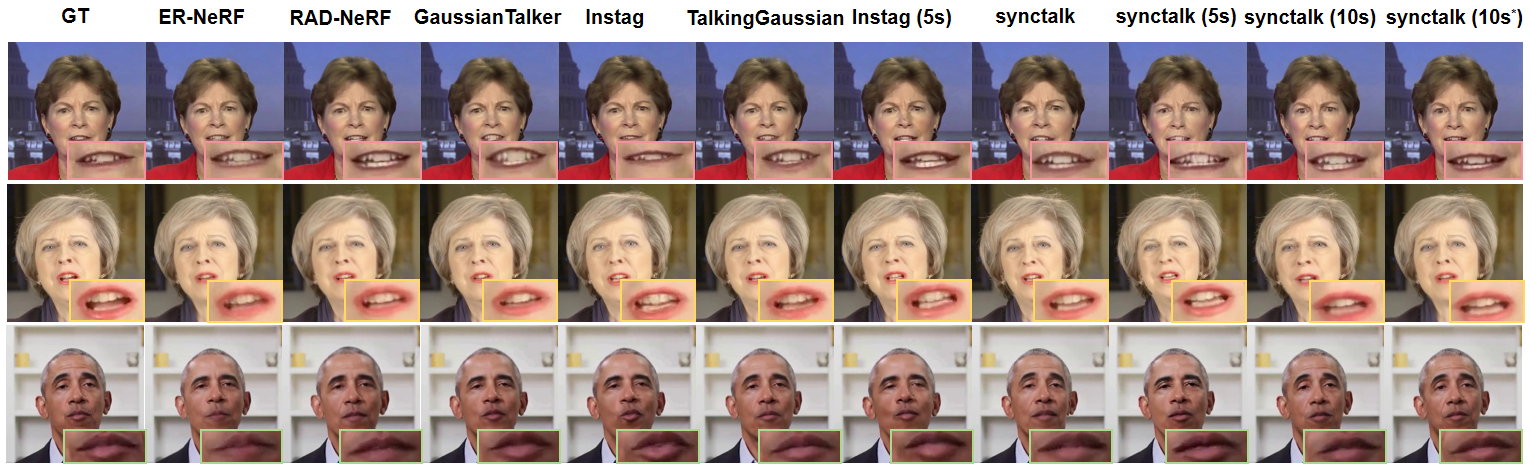}
\caption{Qualitative results on different identities. We compare representative NeRF-based, 3DGS-based, and one-shot methods, as well as our short-segment training strategy. Zoomed-in mouth regions are shown for clearer comparison of lip articulation quality.}
\label{fig:Qualitative_results}
\end{figure*}
\textbf{Qualitative Results}.
Figure~\ref{fig:Qualitative_results} presents qualitative comparisons across representative identities. Several observations can be made. First, 3DGS-based methods generally produce sharper lip boundaries and clearer local textures, while NeRF-based methods often achieve better lip synchronization consistency. Second, ISExplore consistently improves the visual quality of short-segment training for both paradigms. In particular, InstaG with ISExplore exhibits more accurate lip articulation than its random-segment counterpart, while SyncTalk trained on informative 5-second or 10-second segments preserves most of the visual fidelity of full-length training.

These visual results are well aligned with the quantitative findings, showing that ISExplore can effectively preserve facial details and lip motion quality while significantly reducing the amount of required reference data.

\subsection{Ablation and Analysis}

\begin{table}[t]
\centering
\resizebox{0.45\textwidth}{!}{
\begin{tabular}{llll|c|c|c}
\toprule
Audio & Camera & Lip & Top5 & Sync-C $\uparrow$ & Sync-D $\downarrow$ & Processing \\
\midrule
- & - & - & - & 4.97 & 9.51 & 0min \\
\ding{51} & - & - & - & 5.38 & 9.29 & 0.5min \\
- & \ding{51} & - & - & 5.44 & 9.12 & 6min \\
- & - & \ding{51} & - & 5.77 & 8.73 & 6min \\
- & \ding{51} & \ding{51} & - & 5.76 & 8.98 & 6min \\
\ding{51} & \ding{51} & \ding{51} & \ding{51} & 6.00 & 8.97 & 1min \\
\bottomrule
\end{tabular}
}
\caption{Ablation study.}
\label{ab_exp}
\end{table}
\textbf{Ablation Study}.
We conduct ablation experiments on SyncTalk to analyze the contribution of different factors in ISExplore. As shown in Table~\ref{ab_exp}, using only audio diversity already provides a reasonable baseline for identifying useful segments. However, incorporating motion-related factors further improves lip synchronization performance. In particular, combining audio diversity with both lip motion and camera-related motion yields the best overall balance between performance and efficiency.

The results also show that directly extracting 3DMM representations for all segments introduces substantial computational overhead. ISExplore avoids this issue through a coarse-to-fine design: it first performs lightweight filtering based on audio, and then applies motion analysis to only a small subset of candidates. This design achieves a favorable trade-off between efficiency and selection quality.

\begin{table*}[t]
\centering
\caption{User study. Ratings range from 1 to 10, where higher is better. Our method achieves the best time-to-performance ratio.}
\resizebox{\textwidth}{!}{
\begin{tabular}{c|cccc|cc|cc}
\toprule
Methods & ER-NeRF & GaussianTalker & TalkingGaussian & RAD-NeRF & Instag (ours 5s) & Instag (origin) & Synctalk (ours 5s) & Synctalk (origin) \\
 & 3h 52min & 4h 4min & 2h 8min& 4h 20min & 24min & 23min & 29min & 3h 10min \\
\midrule
Lip-sync Accuracy & 5.57 & 5.75 & 6.12 & 5.53 & 5.59 & 5.01 & \underline{6.59} & \textbf{7.2} \\
Image quality & 5.24 & 5.63 & \textbf{6.06} & 5.55 & 5.11 & 4.39 & 5.55 & \underline{5.84} \\
Pose Accuracy & 5.24 & 5.81 & 6.1 & 6.26 & 5.61 & 5.41 & \underline{6.28} & \textbf{7.41} \\
Video realness & 5.11 & 5.48 & \underline{5.88} & 5.66  & 5.22 & 4.48 & 5.64 & \textbf{6.35} \\
Time-to-Performance & 0.09 & 0.09 & 0.19 & 0.08 & \textbf{0.90} & 0.80 & \underline{0.83} & 0.14 \\
\bottomrule
\end{tabular}
}

\label{tab:user_study}
\end{table*}
\textbf{User Study}.
To better evaluate quality in practical human-centered scenarios, we conduct a user study on 72 generated videos from 8 methods trained using 5-second reference segments. Five participants are invited to rate each anonymous method across four aspects: (1) lip-sync accuracy, (2) image quality, (3) pose accuracy, and (4) overall realism.
As shown in Table~\ref{tab:user_study}, ISExplore achieves the best \emph{time-to-performance} ratio. Although some full-reference methods still achieve slightly stronger quality in certain dimensions, they typically require several hours of processing and training time. In contrast, ISExplore enables competitive perceptual quality within approximately 20--30 minutes, demonstrating clear practical advantages for rapid personalization.

\begin{table*}[h!]
\small
\centering
\caption{Additional quantitative validation on 40 extra HDTF subjects, comparing full-length training with the 10-second setting on SyncTalk.}
\label{tab:synctalk_10s}
\resizebox{\linewidth}{!}{
\begin{tabular}{lcccccccccc}
\toprule
\textbf{Method} & \textbf{PSNR}$\uparrow$ & \textbf{MS-SSIM}$\uparrow$ & \textbf{LPIPS}$\downarrow$ & \textbf{FID}$\downarrow$ & \textbf{NIQE}$\downarrow$ & \textbf{BRISQUE}$\downarrow$ & \textbf{Sync-C}$\uparrow$ & \textbf{Sync-D}$\downarrow$ & \textbf{Process Time}$\downarrow$ & \textbf{Training Time}$\downarrow$ \\
\midrule
SyncTalk & 29.65 & 0.9320 & 0.0291 & 6.75 & 23.34 & 34.92 & 8.09 & 7.08 & 1h33min & 2h \\
SyncTalk (10s*) & 28.84 & 0.9265 & 0.0320 & 7.78 & 23.36 & 34.77 & 7.50 & 7.51 & 12.5min($\times$ 7.44) & 36min(x$\times$ 3.3)\\
\midrule
 Ground Truth& $\infty$& 1& 0& 0& 24.52& 48.01& 7.92& 6.95& -&-\\
\bottomrule
\end{tabular}
}
\end{table*}

\begin{table}[t]
\centering
\caption{Expressiveness and extreme condition analysis. Var represents the variance of the camera viewpoint coordinates.}
\label{tab:method_var}
\resizebox{0.75\linewidth}{!}{
\begin{tabular}{c|cccc}
\toprule
\textbf{Methods}
& PSNR$\uparrow$
& Sync-c$\uparrow$
& FID$\downarrow$
& Variance$\uparrow$ \\
\midrule
Synctalk (baseline) & 30.82 & 7.33 & 6.33 & 0.092  \\
Synctalk (10s*) & 30.14 & 6.64 & 5.74 & 0.094  \\
Instag (random) & 28.17 & 5.44 & 9.46 & 0.096\\
Instag (ours) & 29.16 & 6.00 & 9.74 & 0.106\\
\bottomrule
\end{tabular}
}
\end{table}
\textbf{Expressiveness / Diversity Analysis}.
To verify that short reference clips do not compromise motion diversity, we measure the variance of the extracted camera viewpoint coordinates, where a larger variance indicates richer pose dynamics.
As shown in Table~\ref{tab:method_var}, SyncTalk trained with only a 10-second clip maintains viewpoint variance comparable to the full-data baseline (0.094 vs. 0.092), while preserving competitive image quality and fidelity. A similar trend is observed for InstaG: compared with random segment selection, ISExplore improves not only visual quality and lip synchronization, but also motion diversity (0.106 vs. 0.096). These results indicate that compact yet informative clips are sufficient for preserving expressive motion in personalized TFG.

\textbf{Additional Data Efficiency Study}.
To further validate data efficiency, we additionally select 40 random subjects from HDTF. We then compare SyncTalk trained on full-length videos with the 10-second setting.
As shown in Table~\ref{tab:synctalk_10s}, reducing the reference duration from minutes to only 10 seconds causes only limited degradation in both rendering and synchronization quality. For example, PSNR decreases from 29.65 to 28.84, while MS-SSIM, LPIPS, and Sync metrics remain competitive. Notably, the 10-second setting also stays close to the ground-truth reference on several perceptual indicators, suggesting that short-segment training does not significantly distort the natural data distribution. In contrast, the efficiency gain is substantial: processing time is reduced from 1h33min to 12.5 minutes, and training time from 2 hours to 36 minutes. These results reveal strong redundancy in long reference videos and further validate the effectiveness of informative short-segment selection.

%app instag、random消融

\section{Conclusion}
In this paper, we present ISExplore, a simple and effective reference segment selection strategy for efficient personalized talking face generation. By analyzing audio diversity, lip motion, and camera viewpoint variation, ISExplore identifies informative short reference clips for training high-quality TFG models. Extensive experiments show that our method reduces the computational cost of personalized TFG training to under 30 minutes while maintaining competitive visual and motion quality.

\section{Acknowledgments}

\bibliographystyle{ACM-Reference-Format}
\bibliography{sample-base}

\appendix
% \section{Appendix}
\clearpage
% \setcounter{page}{1}
% \maketitlesupplementary

\section{More Experiments}

\subsection{Additional Generalization Study on Extra HDTF Subjects}

To further validate the generalization ability of our method beyond the main benchmark split, we conduct an additional large-scale experiment on 40 randomly selected subjects from the HDTF dataset. 
For each subject, we construct a practical reference setting by truncating the original video to its first 3 minutes, and compare two training strategies under the same duration budget: 
(1) directly using the first 5s or 10s segment of the reference video as a naive short-clip baseline, and 
(2) using the informative segment automatically selected by our ISExplore strategy.

We intentionally compare against the naive \textit{First 5/10} setting rather than full-length training, since the purpose of this experiment is to isolate the effect of \textit{segment selection} under the same reference-duration constraint. 
Specifically, for the 3DGS-based model \textbf{InstaG}, we compare the original \textit{First 5} setting with \textit{Ours 5}. 
For the NeRF-based model \textbf{SyncTalk}, we compare the original \textit{First 10} setting with \textit{Ours 10s}.

\begin{table}[t]
\centering
\caption{Additional quantitative validation on 40 extra HDTF subjects. 
ISExplore consistently outperforms naive short-clip selection on both InstaG and SyncTalk under the same reference-duration budget.}
\label{tab:quantitative_comparison}
\resizebox{\linewidth}{!}{
\begin{tabular}{lcccccc}
\toprule
\textbf{Method} & \textbf{PSNR}$\uparrow$ & \textbf{MS-SSIM}$\uparrow$ & \textbf{LPIPS}$\downarrow$ & \textbf{FID}$\downarrow$ & \textbf{Sync-C}$\uparrow$ & \textbf{Sync-D}$\downarrow$ \\
\midrule
GT & $\infty$ & 1.000 & 0.000 & 0.000 & 7.920 & 6.950 \\
\midrule
InstaG (First 5) & 18.823 & 0.717 & 0.188 & 21.154 & 6.469 & 8.375 \\
InstaG (Ours 5) & \textbf{18.931} & \textbf{0.720} & \textbf{0.186} & \textbf{20.707} & \textbf{7.013} & \textbf{8.063} \\
\midrule
SyncTalk (First 10) & 27.795 & 0.917 & 0.050 & 9.998 & 5.659 & 8.893 \\
SyncTalk (Ours 10s) & \textbf{28.842} & \textbf{0.926} & \textbf{0.032} & \textbf{7.776} & \textbf{7.501} & \textbf{7.514} \\
\bottomrule
\end{tabular}
}
\end{table}

As shown in Table~\ref{tab:quantitative_comparison}, replacing the naively selected initial segment with the informative segment discovered by ISExplore consistently improves performance across both paradigms.

For \textbf{InstaG}, ISExplore improves PSNR from \textbf{18.82} to \textbf{18.93}, MS-SSIM from \textbf{0.7171} to \textbf{0.7200}, and LPIPS from \textbf{0.1884} to \textbf{0.1856}. 
At the same time, it also improves lip-synchronization quality, with Sync-C increasing from \textbf{6.47} to \textbf{7.01} and Sync-D decreasing from \textbf{8.38} to \textbf{8.06}. 
These results indicate that even for an already efficient 3DGS-based pipeline, the quality of the selected short reference segment remains crucial.

For \textbf{SyncTalk}, the gain is even more evident. 
Using the informative 10-second segment selected by ISExplore improves PSNR from \textbf{27.80} to \textbf{28.84}, MS-SSIM from \textbf{0.9168} to \textbf{0.9265}, LPIPS from \textbf{0.0496} to \textbf{0.0320}, and FID from \textbf{10.00} to \textbf{7.78}. 
Meanwhile, lip-synchronization quality is also substantially improved, with Sync-C increasing from \textbf{5.66} to \textbf{7.50} and Sync-D decreasing from \textbf{8.89} to \textbf{7.51}. 
This demonstrates that for NeRF-based personalized TFG, selecting an informative short clip is significantly more effective than simply taking the first available segment.

More importantly, our method is fundamentally \textbf{orthogonal} to prior efforts that reduce data dependency through stronger model priors. 
For example, InstaG already introduces effective 3D priors and efficient representation learning to enable personalized TFG from short videos. 
In contrast, our method focuses on \textit{data selection} rather than \textit{model design}. 
Therefore, the two approaches are inherently complementary rather than competing.

The fact that ISExplore consistently improves \textbf{InstaG} is particularly important, as it demonstrates that informative segment selection remains beneficial even when strong data priors are already incorporated. 
This suggests that our method provides an additional and independent performance gain on top of prior-based acceleration strategies, demonstrating that data-centric selection and prior-based modeling can work synergistically.

Overall, these results further support our central claim: \textbf{high-quality personalized 3D talking face generation can be achieved using only a few seconds of informative reference video}. 
The key challenge is therefore not merely reducing the reference duration, but identifying which short segment is most suitable for training. 
A naive choice such as the first few seconds may contain insufficient articulation, limited viewpoint diversity, or weak audio variation, which can substantially degrade model performance. 
In contrast, ISExplore is able to identify compact yet information-rich reference segments that provide consistently better supervision for both 3DGS- and NeRF-based personalized TFG models.

\subsection{Train scaling}
TTo investigate how the reduction of reference video duration affects the required training budget, we further analyze the impact of training epochs when SyncTalk is trained using only a 5-second reference video. 
For each training length, we select the best checkpoint according to validation performance and evaluate the final output quality using Sync-C and Sync-D.

\begin{figure}[h]
\centering
\includegraphics[width=0.5\textwidth]{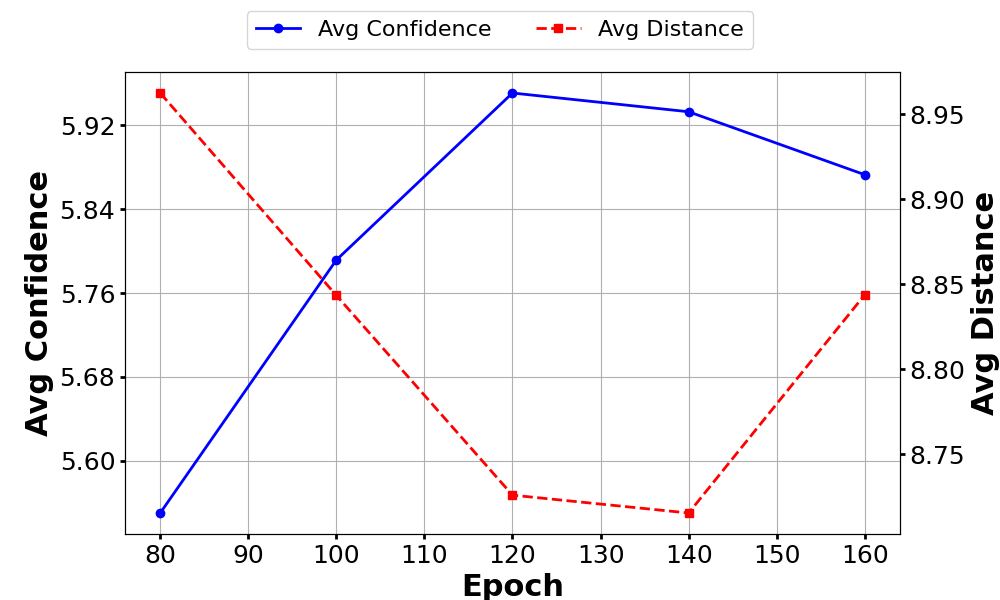}
\caption{Training scaling result on Sync-C and Sync-D metrics.}
\label{training_scale}
\end{figure}

As shown in Fig.~\ref{training_scale}, the model performance first improves and then gradually degrades as the number of training epochs increases. 
Specifically, both lip synchronization confidence and distance metrics indicate that the best performance is achieved at an intermediate training stage rather than at the longest training duration. 
This suggests that, under a highly compressed reference setting, the model does not require the same optimization horizon as conventional long-video training.

A possible explanation is that when only a few seconds of reference data are used, the model can fit the dominant identity, articulation, and pose patterns much earlier. 
Further training may lead to overfitting to the limited reference observations, thereby reducing generalization quality during inference. 
This phenomenon indicates that \textbf{compressing the reference video scale not only reduces data preparation cost, but also shortens the effective training schedule}.

Overall, these results further support our central claim that \textbf{high-quality personalized talking face generation can be achieved with only a few seconds of informative reference content}. 
More importantly, such compact yet informative supervision also enables a substantial reduction in training time, making the overall pipeline significantly more efficient without sacrificing output quality.

\begin{figure}[ht]
\centering
\includegraphics[width=0.5\textwidth]{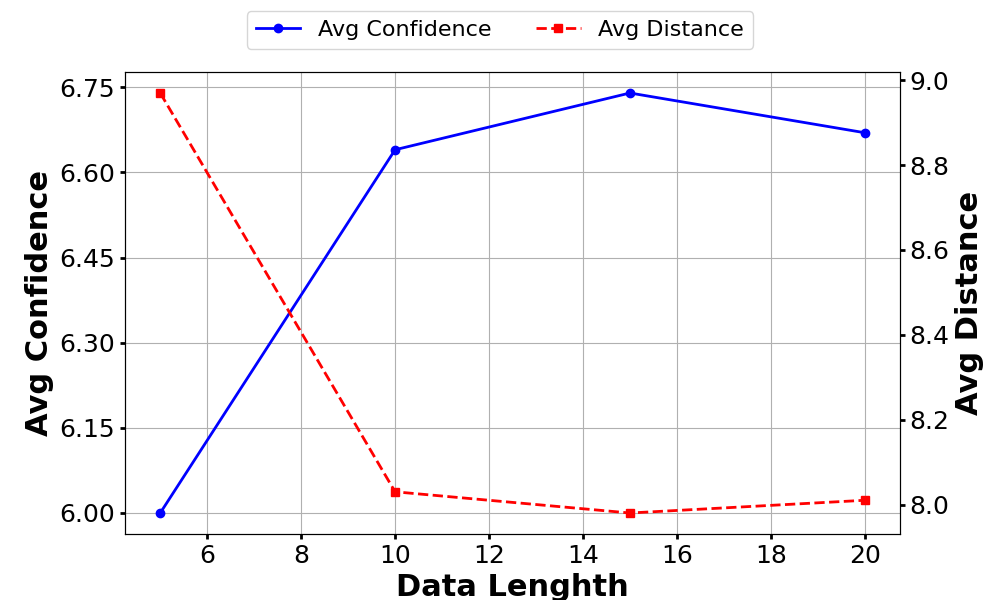}
\caption{Data scaling result on Sync-C and Sync-D metrics.}
\label{data_scale}
\end{figure}

\subsection{Data scaling}
To investigate the impact of reference video length on the quality of the final output, we train SyncTalk using 5-second, 10-second, 15-second, and 20-second reference videos selected by the ISExplore strategy. 
For each setting, the best checkpoint is selected based on validation performance. 
The number of training epochs is fixed to 140 according to the results of the training boundary analysis.

As shown in Fig.~\ref{data_scale}, the output video quality improves significantly when increasing the reference length from 5 seconds to 10 seconds, but becomes relatively stable when further extending the duration to 15 or 20 seconds. 
This indicates that most of the useful information required for high-quality personalized TFG can already be captured within a short time window, and additional data beyond this range provides diminishing returns.

These results further support our central insight that \textbf{high-quality personalized talking face generation does not require long reference videos, but mainly depends on sufficiently informative short segments}. 
Once the key identity, articulation, and motion patterns are captured, extending the reference duration brings limited additional benefit.

Additionally, the consistent performance across different durations demonstrates that ISExplore can effectively identify informative segments under varying reference length settings, making it a robust and flexible data selection strategy.

\subsection{Hyperparameter analysis}
\paragraph{Hyperparameter Analysis.}
\begin{table}[t]
\centering
\caption{Hyperparameter analysis of ISExplore on SyncTalk. We evaluate different segment durations and selection strategies.}
\label{tab:hyperparameter}
\resizebox{\linewidth}{!}{
\begin{tabular}{lcccc}
\toprule
\textbf{Methods} & \textbf{PSNR}$\uparrow$ & \textbf{Sync-C}$\uparrow$ & \textbf{FID}$\downarrow$ & \textbf{ISExplore Processing}$\downarrow$ \\
\midrule
SyncTalk (2s) & 29.04 & 4.21 & 10.74 & 57s \\
SyncTalk (3s) & 29.83 & 4.86 & 12.21 & 58s \\
\midrule
SyncTalk (threshold=0.15) & 30.48 & 6.02 & 10.92 & 1min \\
SyncTalk (threshold=0.35) & 30.30 & 5.37 & 11.87 & 1min \\
\midrule
SyncTalk (top3) & 30.35 & 6.04 & 11.18 & 0min41s \\
SyncTalk (top10) & \textbf{30.70} & \textbf{6.18} & \textbf{7.50} & 1min32s \\
\midrule
SyncTalk (threshold=0.25, top=5, 5s) & 30.51 & 6.00 & 8.11 & 1min \\
\bottomrule
\end{tabular}
}
\end{table}
We further analyze the impact of key hyperparameters in ISExplore, including segment duration, threshold selection, and candidate pool size. 
As shown in Table~\ref{tab:hyperparameter}, several important observations can be made.

First, \textbf{segment duration plays a critical role in providing sufficient training information}. 
When the duration is reduced to 2s or 3s, the model performance degrades significantly in both rendering quality and lip synchronization. 
This indicates that although our method aims to minimize data usage, excessively short segments may lack sufficient articulation and audio diversity for stable learning. 
In contrast, slightly longer segments (e.g., 5s) provide a better balance between compactness and informativeness.

Second, \textbf{the threshold for motion complexity significantly affects segment quality}. 
A lower threshold (e.g., 0.15) tends to retain segments with richer motion and audio variation, leading to better synchronization performance, while a higher threshold (e.g., 0.35) may over-filter informative content and degrade performance. 
This validates our design choice of balancing motion complexity and information richness.

Third, \textbf{the candidate pool size (top-$k$) introduces a trade-off between performance and efficiency}. 
Using a larger candidate set (e.g., top-10) leads to the best overall performance, achieving the highest PSNR and Sync-C as well as the lowest FID, at the cost of slightly increased processing time. 
In contrast, smaller candidate sets (e.g., top-3) are more efficient but may miss the most informative segments.

Finally, combining these factors, the setting of \textbf{threshold=0.25, top-5} achieves a favorable trade-off between performance and efficiency, which is adopted as the default configuration in our experiments.

Overall, these results further support our core insight: \textbf{high-quality personalized TFG does not require long reference videos, but relies on selecting sufficiently informative short segments}. 
Proper hyperparameter choices enable ISExplore to effectively identify such segments, achieving strong performance while maintaining low computational cost.
\section{More Implementation Details}

\subsection{Implementation Details for Preliminary Analysis}

This section provides implementation details for the controlled preliminary studies in Section~3, where we analyze how different properties of short reference segments affect the final performance of personalized talking face generation (TFG) models.

\paragraph{Base Model and Training Setup.}
We adopt the official implementation of SyncTalk~\cite{Peng_2024_synctalk} as the main analysis platform, since it showed the strongest lip synchronization and head pose rendering quality among the representative NeRF-based personalized TFG methods in our preliminary tests. 
Following the official setup, we use the AVE module proposed in SyncTalk as the audio feature encoder.

\paragraph{Controlled Setup for Audio Diversity Analysis.}
To analyze the impact of audio feature diversity in isolation, we select a 4-minute reference video (\textit{May}) from the GeneFace dataset~\cite{Ye_2024_geneface}. 
We train a SyncTalk model on the full reference video and use the original reference audio as the driving input during inference.

To decouple the influence of head motion from audio-driven lip dynamics, we fix the camera pose during inference, resulting in a static-head output video. 
This controlled setting allows us to focus on the effect of audio-related factors without interference from viewpoint variation. 
The generated 240-second output video is then divided into 48 non-overlapping 5-second segments for efficient evaluation. 
We use non-overlapping segmentation to reduce redundancy and computational cost while still covering diverse temporal content. 
Since training a SyncTalk model from scratch typically requires more than 2 hours, the total runtime of this controlled study exceeds 100 hours.

\paragraph{Lip Motion Complexity Analysis.}
To analyze the impact of lip motion complexity on model performance, we extract frame-wise 3DMM lip landmarks (points 48--68) from the 48 generated segments using the method of \cite{Bulat_2017_3DMM_extracter}. 
These landmarks are used to construct temporal lip motion trajectories, which are further transformed into the frequency domain for High-Frequency Ratio (HFR) analysis as described in Section~3.4.

\paragraph{Controlled Setup for Camera View Analysis.}
To analyze the effect of camera-view complexity, we simulate reference videos with identical audio content but different camera pose trajectories. 
Specifically, we re-render reference videos under controlled camera settings such that the number of distinct viewpoints within a segment is fixed to 5, 10, 20, or 50, respectively. 
Following the SyncTalk pipeline, when the available pose sequence is shorter than the target duration, we apply mirrored extension to complete the camera trajectory. 
Each simulated reference video is then used to train a SyncTalk model from scratch under the same training configuration.

\paragraph{Evaluation Metric.}
For the controlled analyses in Section~3, we primarily use Sync-C~\cite{Prajwal_2020_sync_c_d} as the evaluation metric to measure lip synchronization quality. 
This is because radiance field-based personalized TFG models can reconstruct static facial regions (e.g., forehead and cheeks) with high fidelity regardless of the audio input~\cite{Athar_2022_RigNeRF,Gu_2022_styleNerf}. 
As a result, image-level metrics such as PSNR or SSIM are less sensitive to the specific factors investigated in these preliminary studies. 
In contrast, Sync-C provides a more direct reflection of how different reference segment properties influence the learned audio-to-lip correspondence.

\subsection{Baseline Training Consistency.}
All baseline models are trained using the official implementations released by the corresponding authors. 
When AVE is supported as the audio encoder in the official codebase, we adopt it for consistency; otherwise, we follow the original audio encoding and training settings specified in the respective papers.
\end{document}